%% file: main.tex
\documentclass[sigconf]{acmart}

\usepackage{times}
\usepackage{latexsym}
\usepackage{url}
\usepackage{wasysym}
\usepackage{comment}
\usepackage{placeins}
\usepackage{array,multirow,graphicx}
\usepackage{float}
\usepackage{hyperref}

\newcommand{\BiLSTM}{BiLSTM }
\newcommand{\BiCLSTM}{BiCLSTM }
\newcommand{\BiLSTMTI}{TACAM-WE }
\newcommand{\BiLSTMKG}{TACAM-KG }
\newcommand{\Bert}{CAM-BERT Base}
\newcommand{\BertTI}{TACAM-BERT Base }
\newcommand{\BertLarge}{CAM-BERT Large }
\newcommand{\BertLargeTI}{TACAM-BERT Large }

\AtBeginDocument{%
  \providecommand\BibTeX{{%
    \normalfont B\kern-0.5em{\scshape i\kern-0.25em b}\kern-0.8em\TeX}}}

\setcopyright{acmcopyright}
\copyrightyear{2019} 
\acmYear{2019} 
\acmConference[WI '19]{IEEE/WIC/ACM International Conference on Web Intelligence}{October 14--17, 2019}{Thessaloniki, Greece}
\acmBooktitle{IEEE/WIC/ACM International Conference on Web Intelligence (WI '19), October 14--17, 2019, Thessaloniki, Greece}
\acmPrice{15.00}
\acmDOI{10.1145/3350546.3352506}
\acmISBN{978-1-4503-6934-3/19/10}

\begin{document}

\title{TACAM: Topic And Context Aware Argument Mining}

\author{Michael Fromm}
\email{fromm@dbs.ifi.lmu.de}
\affiliation{%
  \institution{Database Systems and Data Mining, LMU}
  \city{Munich}
  \country{Germany}
}

\author{Evgeniy Faerman}
\email{faerman@dbs.ifi.lmu.de}
\affiliation{%
  \institution{Database Systems and Data Mining, LMU}
  \city{Munich}
  \country{Germany}
}

\author{Thomas Seidl}
\email{seidl@dbs.ifi.lmu.de}
\affiliation{%
  \institution{Database Systems and Data Mining, LMU}
  \city{Munich}
  \country{Germany}
}

\renewcommand{\shortauthors}{Fromm and Faerman, et al.}

\begin{abstract}
In this work we address the problem of argument search. The purpose of argument search is the distillation of pro and contra arguments for requested topics from large text corpora. In previous works, the usual approach is to use a standard search engine to extract text parts which are relevant to the given topic and subsequently use an argument recognition algorithm to select arguments from them. The main challenge in the argument recognition task, which is also known as argument mining, is that often sentences containing arguments are structurally similar to purely informative sentences without any stance about the topic. In fact, they only differ semantically. Most approaches use topic or search term information only for the first search step and therefore assume that arguments can be classified independently of a topic. We argue that topic information is crucial for argument mining, since the topic defines the semantic context of an argument. Precisely, we propose different models for the classification of arguments, which take information about a topic of an argument into account. Moreover, to enrich the context of a topic and to let models understand the context of the potential argument better, we integrate information from different external sources such as Knowledge Graphs or pre-trained NLP models. Our evaluation shows that considering topic information, especially in connection with external information, provides a significant performance boost for the argument mining task.
\end{abstract}





\maketitle

\input{intro}

\input{related_work}
\input{method}
\input{evaluation}

\input{conclusion.tex}

\begin{acks}
This work has been funded by the Deutsche Forschungsgemeinschaft (DFG) within the project Relational Machine Learning for Argument Validation (ReMLAV), Grant Number SE 1039/10-1, as part of the Priority Program "Robust Argumentation Machines (RATIO)" (SPP-1999). This work has also been funded by the German Federal Ministry of Education and Research (BMBF) under Grant No. 01IS18036A. The authors of this work take full responsibilities for its content.
\end{acks}

\bibliographystyle{ACM-Reference-Format}
\bibliography{main}

\end{document}

%% file: intro.tex
\section{Introduction}
\begin{figure}
  \centering
    \includegraphics[width=0.35\textwidth]{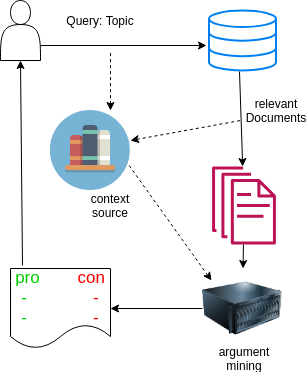}
    \caption{Argument search pipeline with context}
    \label{fig:overview}
\end{figure}
The main focus of argument search lies on presenting an overview of different standpoints and their justifications to some inquired topic e.g. \textit{cloning} or \textit{minimum wages}. This may be useful in different scenarios, like legal reasoning \cite{Wyner2010} or decision making processes \cite{SVENSON197986}, especially if a topic or a problem is controversial.
An automated argument search process could ease much of the manual effort involved in these areas, especially if it can make use of large text databases or even combinations of them. 
The online argument search in state-of-the-art argument search systems proceeds in two steps \cite{N18-5005}:
\begin{enumerate}
\item Some standard text search engine, e.g. \cite{bialecki2012apache}, extracts relevant text parts from large text corpora using a given topic as a query.
\item Relevant text parts are analyzed sentence-wise by an argument recognition component which decides for each sentence whether it is an argument and optional about its stance.
\end{enumerate}
Therefore, the core technique in argument search is argument recognition or argument mining \cite{RePEc:igg:jcini0:v:7:y:2013:i:1:p:1-31, Lippi:2015:CCD:2832249.2832275, swanson-etal-2015-argument, Villalba2012SomeFO, 10.1007/978-3-319-28460-6_10}.
The basis for argument mining is an argument model (to avoid confusion with machine learning models in the following we refer to argument model as argument scheme). An argument scheme formally defines what kind of arguments exist and what their properties and relationships between them are. State-of-the-art argument search systems work with simple argument models without relationships between arguments. Therefore, the common task of a machine learning model is argument recognition or identification.

The classic argument recognition approaches extract arguments from text without taking the topic of the argument into consideration \cite{ Palau:2009:AMD:1568234.1568246, DBLP:journals/corr/HabernalG16, DBLP:journals/corr/EgerDG17}. However, the special characteristic of application of argument recognition in argument search is that there is always a query topic. The query topic carries information about the query context and understanding the context of potential arguments can be crucial for the decision. For instance, if the query is about the usefulness of some medical procedure in the context of medicine, we expect appropriate arguments from the medical doctors and not from people who share their own individual experiences. Thus, if a potential argument follows a particular structure or some special terminology is used, it may increase an argument's chances to be classified as an argument. 

Another desirable property of an argument identification approach is to be able to decide \textbf{dependent} on a topic, whether a sentence is an argument. 
To be useful, an argument search engine heavily relies on a large text corpus. The larger the text corpus is, the more probable is the scenario that texts extracted by the search engine contain arguments about different topics, which increases coverage of different topics but further complicates the argument classification. For instance, consider the following example: A user looks for arguments to \textit{Emission Trading System (ETS)} and the following sentence candidates are retrieved by the text search engine:
\begin{itemize}
    \item ETS sets a clear price on carbon and combats climate change. 
    \item Free trade secures all the advantages of international division of labour.
    \item UK signals plan to leave EU emissions trading scheme after Brexit.
\end{itemize}
The first two sentences can be considered as arguments, the third sentence is purely informative and it does not persuade towards any stance. However, if we look more closely at the second sentence, we recognize that this is not an argument about the query topic \textit{Emission Trading}. This is obvious for humans, since we understand that the context of \textit{Free Trade} is different from the \textit{Emission Trading} context, even if both contexts are related to \textit{Trade}. Therefore, the better the machine learning model is able to grasp the context of a topic and of potential arguments at different granularities, the better is the decision the model can make and the more certain it can be about its decisions.
Considering relationship between potential argument and topic is different from the classical relation detection task in argument mining.
The input to relation detection algorithm are parts of the text, which are already recognized as arguments and presence or absence of some relationships does\textbf{} not affect decision about text parts of being argumentative. In contrast, our approach takes the relationship to the topic into account  
when deciding whether a text is an argument.

In this work we propose a new approach for argument mining which also takes the topic of potential arguments into account. The overview of our approach is depicted in Figure \ref{fig:overview}. The standard argument search pipeline looks like the workflow presented in the figure without the context source and the dotted arrows. Our approach enriches the argument candidates with the context and topic information in the classification process.
We show how the contextual information about a topic and an argument from different sources like knowledge graphs or pre-trained models can be integrated into our approach.
We investigate the benefits of considering the topic and the integration of external knowledge. We summarize our main contributions as follows:




\begin{itemize}
    \item We present a novel approach for argument classification which takes the topic of the argument into account by extending the methodology introduced by the authors of \cite{darmstadt_1}.
    \item We show how contextual information about topic and argument from different sources like knowledge graphs or pre-trained models can be integrated.
    \item We demonstrate that considering topics is beneficial for the argument classification, especially in connection with external knowledge. 
    \item We show that our approach is particularly successful if the model has to generalize to unseen topics. 
    Since we cannot expect that available training datasets for argument recognition cover all possible topics, the generalization to unseen topics is an important requirement.
    \item We present thorough experimental evaluations of our models and comparisons to state-of-the-art methods on a real-world dataset and introduce an additional experimental setting. In this setting we evaluate the ability of different models to classify in the context of topics.
\end{itemize}







%% file: related_work.tex
\section{Related Work}
In general, the main focus in argument mining lies in the recognition of argument components \cite{Palau:2009:AMD:1568234.1568246, DBLP:journals/corr/HabernalG16, stab-gurevych-2017-parsing, article2,nguyen-litman-2015-extracting} and the detection of relations between them \cite{stab-gurevych-2017-parsing, nguyen-litman-2016-context}.
However, all these approaches which tackle the problem of argument classification do not take information about the specific topic of a given argument into consideration.

At the same time different argumentation schemes of different complexity were proposed in previous works \cite{Toulmin1958-TOUTUO-2, article, Freeman2011-FREASR-2, DBLP:conf/emnlp/StabG14}. Since each argumentation scheme contains different numbers of various argument types, this has an implication on machine learning models designed for argument detection, since they have to learn how to identify them.
However, as was shown in
\cite{DBLP:journals/corr/DaxenbergerEHSG17},  these argumentation schemes do not generalize well to different types of texts.
Concretely, the authors of this work collected datasets used with different argumentation schemes and combined them in a single dataset. Afterwards, they trained a model, which should detect the argument component of type \textit{claim}, which is central in each argument scheme. However, the machine learning models which perform well for single datasets could not achieve good results on this simple binary classification task. Additionally, it was shown that even human annotators
often label differently when annotating the same datasets according to complex argumentation schemes. Therefore, the authors came to the conclusion that certain argument components (backing, warrant) as introduced in \cite{Toulmin1958-TOUTUO-2}, and other argumentation schemes are often only stated implicitly in common argumentation documents on the internet.
In more recent work, argumentation schemes became simpler and more flexible \cite{darmstadt_1, wachsmuth-etal-2017-building}. 
This enables broader applicability and topic-dependent argument search across multiple text types. 

There are various approaches to consider context in argument mining. Hand-crafted features extracted from source text were used for argument classification \cite{nguyen-litman-2015-extracting} and relation detection \cite{nguyen-litman-2016-context}. More related to our work is a method presented in \cite{darmstadt_1}. The authors introduced a dataset with arguments of different text types and topics for each argument. Additionally, they propose two simple argumentation schemes. The first scheme is a binary decision, aiming at classifying a sentence as argumentative or non-argumentative. In the second scheme there is a distinction between non argumentative sentences and pro and contra arguments. They also propose a model which takes topics into consideration. We extend their work by proposing new architectures and context sources and compare our approach with their method.

There are few approaches which use transfer learning for the argument mining task. In \cite{darmstadt_1} the proposed model is pre-trained on another dataset for argument mining \cite{Habernal.et.al.SIGIR.2016}, but this approach does not lead to considerable improvement.
Parallel to our work, the authors of \cite{DBLP:journals/corr/abs-1904-09688} also use transfer learning with BERT for a new introduced corpus with tagged sequences. However, their model does not generalize to the new topics by design.

Based on recent developments two argument search engines, i.e., \emph{\url{www.args.me}} \cite{wachsmuth-etal-2017-building} and \emph{\url{www.argumentsearch.com}}  \cite{stab-etal-2018-argumentext}, where a user is able to search a broad range of documents for certain topics, have been developed.

%% file: method.tex
\section{Problem setting}
We model the recognition of argumentative sentences as a classification task. Given a sentence $s = \{s_{0}, \dots, s_{n}\}$ and a topic $t = \{t_{0}, \dots, t_{k}\}$ with $s_{i} \in \{0,1\}^V$, $t_{i} \in \{0,1\}^V$ being one-hot encoded vectors, and $V$ being the size of the vocabulary, we seek to classify $s$ as "contra argument" or "pro argument" if the sentence $s$ includes evidence for supporting or opposing the topic $t$. If the sentence does not contain evidence, it is classified as a "non-argument".  
\section{Method}
In contrast to previous approaches, we aim at incorporating context information into the learning procedure when training our models. This way, the models learn which argument properties are especially meaningful in the context of a particular topic and can put a special emphasis on these information for the subsequent classification task. For instance, \emph{emission trading} is a frequently discussed topic, but we would expect the most meaningful arguments about its usefulness coming from particular academic communities. Consequently, by providing topic information in a meaningful way, we enable models e.g. to learn argument structures and vocabulary which are common in those communities. On the other hand we also expect our models to learn how topics are related to their domain specific arguments. Although a sentence might contain topic-specific words it may still be an argument of a different topic. Considering the topic \textit{emission trading} again, relevant arguments are probably more related to climate change than to the stock market, though \emph{trading} is a frequently used term in the latter area. Thus, it is important to \textit{understand} the context of the topic and the context of the potential arguments. Consequently, we propose various approaches to provide context information about topic and potential argument from various external sources. However, as the proposed models should be able to generalize to arbitrary topics, we provide the context information as an additional input to the models. Therefore, all our models aggregate the representation of the potential argument with the representation of the topic.

\subsection{Models}

\subsubsection{Recurrent Network}
\label{subsubsection:Recurrent_Network}
The first model we propose is a recurrent model for which we use two instances of a \BiLSTM \cite{Hochreiter:1997:LSM:1246443.1246450} model. Precisely, one is used to encode a topic and the other model aims at encoding the potential argument:
\begin{center}
\begin{eqnarray*}
    x^{s} &=& \{s_{1}W^{we}, \dots, s_{n}W^{we}\}\\ 
    h^s &=& BiLSTM_{a}(x^{s})\\\\
    x^{t} &=& f_{map}(t)\\
    x^{t} &=& \{x^{t}_{1}W^{te}, \dots, x^{t}_{m}W^{te}\}\\
    h^t &=& BiLSTM_{t}(x^{t})\\\\
    h_l &=& aggr(h^s, h^t)\\
    \hat{y} &=& softmax(h_{l}W_{final}+b_{final})
\end{eqnarray*}
\end{center}
We use \emph{word2vec} \cite{mikolov2013efficient} embeddings $W^{we} \in \mathbb{R}^{V\times d}$ of the given words in a sentence $s$ as input for the argument \BiLSTM instance $BiLSTM_a$. However, it is noteworthy that any other kind of word embeddings can be used, too. 
Furthermore, function $f_{map}$ maps some given topic description $t$ to a sequence of entities $x^t$. In general, we allow arbitrary information sources to provide topic context. Therefore, $f_{map}$ depends on the information source. In case of describing the relevant entities of $t$ in terms of relevant words, one could use a sequence of word embeddings to encode the topic information. In this case $f_{map}$ would map the relevant words to the corresponding one-hot encoded vectors which, if multiplied with the word embedding matrix $W^{we}$, serve as input for the topic BiLSTM instance denoted as $BiLSTM_{t}$. In case of using knowledge graphs as external source of information for the context, $f_{map}$ first examines whether there is an entity with the same name as the whole  topic description. Otherwise it maps each word in the topic description to an corresponding entity in the knowledge graph. If there is no such corresponding entity for a particular word, we employ a nearest neighbor search for this word in the word embedding space and finally use a knowledge graph entity which matches to a semantically similar word. Once we found an entity for each word in the topic description, we use the corresponding sequence of knowledge graph entity representations as input for the topic BiLSTM instance.
The function $aggr$ is used to aggregate topic and argument representations. We evaluate the following aggregation functions:
\begin{itemize}
    \item Addition: $aggr(h^s, h^t) = h^s + h^t$
    \item Hadamard product: $aggr(h^s, h^t) = h^s \odot h^t$
    \item Concatenation: $aggr(h^s, h^t) = concat(h^s, h^t)$
\end{itemize}
Finally, we use the aggregated representation $h_{l}$ as input to a dense layer with softmax activation to obtain the classification result $\hat{y}$.

\subsubsection{Attention model}
\label{subsubsection:attention_model}
\label{subsub:attention_mode}
We also use a deep bidirectional transformer encoder \cite{vaswani2017attention}, the architecture which was used in BERT \cite{bert}. Specifically, we concatenate argument and topic description and use a special separator token and segment embeddings to distinguish between topic and potential argument. The output of the first special [CLS] token is used as input to the dense classification layer, which predicts the distribution over the classes.

\subsection{Context source}\label{context}
As mentioned previously, our models are able to rely on different external sources that may provide the context information. In this work, we experiment with the following sources:
\begin{itemize}
    \item Shallow \textbf{Word Embeddings} \cite{mikolov2013efficient,pennington2014glove,bojanowski2017enriching} are widely used in NLP applications and encode context information implicitly. In fact, the word embeddings are learned such that the representations of words that frequently appear in similar contexts are similar to each other. We use shallow word embeddings trained by word2vec as input to the recurrent model.
    \item \textbf{Knowledge Graphs} model information about the world explicitly in the form of an heterogeneous graph. The entities in the knowledge graph are represented as nodes, and relationships between them as edges of different types. Information in a knowledge graph is represented as triples consisting of subject, predicate and object, where subject and object are entities and predicate stands for the relationship between them. In contrast to information contained in text data, knowledge graphs are structured, i.e., each entity and relationship have a distinct meaning, and the information about the modelled world are distilled in form of facts. These facts can be extracted from texts, different databases or inserted manually. The trustworthiness of these facts in publicly available knowledge graphs is in general very high \cite{nickel2015review}.
    In our work we use the english version of the DBpedia knowledge graph, which has about 400 million facts  with more than 3.7 million unique entities \cite{dbpedia-swj}. We applied TransE \cite{NIPS2013_5071} to obtain embeddings for the knowledge graph entities. These embeddings are used as input to a recurrent model (alternatively to the word embeddings).
    \item Fine-Tuning based \textbf{Transfer Learning} approaches \cite{radford2018improving,bert,radford2019language} adapt whole models, that were pre-trained on some (auxiliary) task, to a new problem. This is different from feature-based approaches which provide pre-trained representations \cite{Peters:2018,dai2015semi} and require task-specific architecture for a new problem. We use the weights of pre-trained BERT (Large and Base) \cite{bert} models for initializing our \ref{subsub:attention_mode} model and train it for the argument classification task. 
\end{itemize}





%% file: evaluation.tex
\section{Evaluation}
\subsection{Dataset and Evaluation Tasks}
For the evaluation we use the UKP Sentential Argument Mining corpus from \cite{darmstadt_1}. The dataset consists of more than 25000 sentences from multiple text types covering eight different topics. It contains a broad range of genres including news reports, editorials, blogs, debate forums and encyclopedia articles which are all related to at least one topic. The topics have been randomly selected from a list\footnote{\url{https://www.questia.com/library/controversial-topics}} of controversial topics.  
The authors define an argument as a sentence that can be used to oppose or support a given topic.
For all models each sentence is truncated to 60 words according to the experiment setting in \cite{darmstadt_1}. Note that in contrast to \cite{darmstadt_1} we use weighted cross-entropy to account for class imbalance.\footnote{We assume this is a reason we obtained better results for the comparison methods as stated in the original paper.}
Following \cite{darmstadt_1} we evaluate our approach by performing the following classification tasks:
\begin{itemize}
    \item Binary classification: whether a sentence is an argument for the given topic.
    \item Multiclass classification: whether a sentence is supporting, respectively attacking an argument, or is not an argument at all for the given topic.
\end{itemize}

As suggested in \cite{darmstadt_1}, we evaluate all approaches in two different scenarios. In the \textit{In-Topic} scenario each topic is split into training and test data, which leads to arguments of the same topics in both training and test data. The \textit{Cross-Topic} scenario primarily aims at evaluating the generalization of the models, i.e., answering the question how good the performance of the models is on yet unseen topics. Therefore, seven topics are used for training and the remaining one for test. Let us mention that although \textit{Cross-Topic} is the more complex task, it is more relevant for real-world problems: The reason is that in general we cannot expect all possible topic queries to be present in a dataset that is available for training.

\subsection{Models}
For all tasks we compare the following approaches:
\begin{itemize}
    \item \BiLSTM is a bidirectional LSTM model \cite{Hochreiter:1997:LSM:1246443.1246450}, which does not use topic information 
    \item \BiCLSTM is the contextual biderectional LSTM \cite{DBLP:journals/corr/GhoshVSRDH16}. Topic information is used as an additional input to the gates of an LSTM cell. We use the version from \cite{darmstadt_1} where the topic information is only used at the $i-$ and $c-$gates since this model showed the most promising result in their work.
    \item \BiLSTMTI is our recurrent model described in \ref{subsubsection:Recurrent_Network} which uses word embeddings to define the context of the topic
    \item \BiLSTMKG is our recurrent model described in \ref{subsubsection:Recurrent_Network} which uses Knowledge Graphs embeddings from DBPedia to define the context of the topic.
    \item \BertTI $/$ \BertLargeTI are our attention based models with topic information described in Section \ref{subsub:attention_mode}. Both model use pre-initialized weights (cf. Section \ref{context}). \BertTI has 1/3 parameters of \BertLargeTI.
    \item \Bert $/$ \BertLarge are similar to \BertTI and \BertLargeTI models without topic information. These models enrich only potential argument with the context, but do not have access to the topic. Comparing them with their counterparts with topic information enables the evaluation of topic importance.
\end{itemize}

In our experimental setting we mostly follow the experimental settings suggested in \cite{darmstadt_1}. We use the same train/validation/test splits. The validation set is used to select the hyperparameters and we report Macro F1 scores on test sets. To avoid effects of bad initialization and local minima we train each model 10 times and select the model which performs best on the validation set.

\subsection{In-topic Results}
The results of the in-topic argument classification are listed in Table \ref{it}. In this setting we do not expect a large improvement by providing topic information since the models have already been trained with arguments of the same topics as in the training set. The results in Table \ref{it} reflect our expectations: we can slightly improve the classification results for the more complex multiclass classification problem. However, we see a relative increase of about 10\% for the two-classes and 20\% for the three-classes classification problem by using context information from transfer learning. Therefore, we conclude that contextual information about potential arguments is important and since the topics are diverse, the model is able to learn argument structure for each topic.

\begin{table}[htb]
\begin{center}
\begin{tabular}{|l|l|l}
\cline{2-2}
\multicolumn{1}{l|}{}& Method& \\
\hline
\parbox[t]{2mm}{\multirow{6}{*}{\rotatebox[origin=c]{90}{two-class}}}
&\BiLSTM &  \multicolumn{1}{l|}  {0.74}\\ 
&\BiCLSTM&  \multicolumn{1}{l|}{0.74}  \\
&\BiLSTMTI& \multicolumn{1}{l|}{0.74} \\
&\BiLSTMKG& \multicolumn{1}{l|}{0.73} \\
&\Bert & \multicolumn{1}{l|}{0.79} \\
&\BertTI &   \multicolumn{1}{l|}{\bf0.81} \\
&\BertLarge&\multicolumn{1}{l|}{ 0.80} \\
&\BertLargeTI&\multicolumn{1}{l|}{\bf 0.81}\\
\hline
\parbox[t]{2mm}{\multirow{6}{*}{\rotatebox[origin=c]{90}{three-class}}} 
&\BiLSTM  & \multicolumn{1}{l|}{0.56} \\
&\BiCLSTM  & \multicolumn{1}{l|}{0.53} \\
&\BiLSTMTI  & \multicolumn{1}{l|}{0.54} \\
&\BiLSTMKG  & \multicolumn{1}{l|}{0.56} \\
&\Bert 	& \multicolumn{1}{l|}{0.65} \\
&\BertTI&  \multicolumn{1}{l|}{0.66} \\
&\BertLarge & \multicolumn{1}{l|}{0.67} \\
&\BertLargeTI&  \multicolumn{1}{l|}{\bf 0.69} \\

\hline
\end{tabular}
\end{center}
\caption{\label{it} In-Topic}
\end{table}


\subsection{Cross-Topic Results}
\begin{table*}
\begin{center}
\begin{tabular}{|c|c|c|c|c|c|c|c|c|c|c|}
\cline{2-10}
\multicolumn{1}{c}{} & \multicolumn{1}{|c}{Method} & \multicolumn{8}{|c|}{Topics} \\
\cline{2-11} \multicolumn{1}{c}{} & \multicolumn{1}{|c|}{} & \multicolumn{1}{c|}{Abortion} & \multicolumn{1}{c|}{Cloning} &\multicolumn{1}{c|}{Death penalty} & \multicolumn{1}{c|}{Gun control }& \multicolumn{1}{c|}{Marij. legal.} & \multicolumn{1}{c|}{Min. wage} & \multicolumn{1}{c|}{Nucl. energy} &  \multicolumn{1}{c|}{School unif.} & \multicolumn{1}{c|}{$\diameter$}\\ \hline
 \parbox[t]{2mm}{\multirow{6}{*}{\rotatebox[origin=c]{90}{two-classes }}} & \BiLSTM & 0.61 & 0.72 & 0.70 & 0.75 & 0.64 & 0.62 &	0.67 & 0.54 & 0.66 \\
& \BiCLSTM & 0.67 & 0.71 & 0.71 & 0.73 & 0.69 &	0.75 & 0.71 & 0.58	& 0.70 \\
& \BiLSTMTI & 0.64 & 0.71 &	0.70 & 0.74 & 0.64 & 0.63 & 0.68 & 0.55	& 0.66 \\
& \BiLSTMKG & 0.62 & 0.69 &	0.70 & 0.75 & 0.64 & 0.76 & 0.71 & 0.56 &	0.68\\
& \Bert & 0.61 & 0.77 &	0.74 & 0.76 & 0.74 & 0.61 & 0.76 & 0.73 &	0.72 \\
& \BertLarge & 0.62 &	0.79 & 0.75 &	0.77 &	0.77 & 	0.65 & 0.75 &	0.73 &	0.73 \\
& \BertTI & 0.78 & 0.77 & \bf 0.78 & 0.80 &	\bf0.79 &	0.83 &	0.80 & \bf0.83 &	0.80 \\
& \BertLargeTI & \bf 0.79 &	\bf0.78 & \bf0.78 &  \bf0.81 &	\bf0.79 & \bf0.84 & \bf0.83&0.82 &	\bf0.80 \\
 \hline
 \parbox[t]{2mm}{\multirow{6}{*}{\rotatebox[origin=c]{90}{three-classes }}}&  \BiLSTM & 0.47 & 0.52 &	0.48 & 0.48 & 0.44 & 0.42 & 0.48 & 0.42 & 0.46 \\
& \BiCLSTM & 0.49 &	0.52 &	0.46 &	0.51 &	0.46 &	0.44 &	0.47 &	0.42 &	0.47\\
& \BiLSTMTI & 0.47 & 0.52 &	0.47 &	0.48 &	0.46 &	0.46 &	0.48 &	0.41 &	0,47\\
& \BiLSTMKG & 0.46 & 0.51 &	0.47 &	0.47 & 0.46 & 0.48 & 0.47 &	0.41	& 0.47\\
& \Bert & 0.38 & 0.63 &	0.53 & 0.49 & 0.54 & 0.54 &	0.61 & 0,50 &	0.53\\
& \BertTI & 0.42 &	0.68 &	0.54 &	0.50 &	0.60 &	0.49 &	0.64 & 	\bf 0.69 &	0.57\\
& \BertLarge & 0.53 &	0.67 &	0.56 &	0.53 &	0.59 &	0.66 &	0.67 &	0.66 & 0.61\\
& \BertLargeTI & \bf 0.54 & \bf 0.69 &	\bf0.59 &	\bf0.55 &	\bf0.63 &	\bf0.69 &	\bf0.71 &	\bf0.69 &	\bf0.64\\
\hline
\end{tabular}
\end{center}
\caption{\label{ct-3} Cross-Topic}
\end{table*}

Our cross-topic results are presented in Table \ref{ct-3}.
In this experiment, which reflects a real-life argument search scenario, we want to prove our two hypotheses:
\begin{itemize}
    \item When classifying potential arguments, it is advantageous to take information about the topic into account. 
    \item The context of an argument and topic context are important for the classification decision.
\end{itemize}
On the whole, we can see that our two hypotheses are confirmed. In the two-classes scenario the recurrent model improves if topic information is provided by knowledge graph embeddings. By using attention-based models with pre-trained weights we can observe a significant performance boost of eleven score points in average when considering topic information. However, the same model without topic information performs only slightly better than the recurrent models. Therefore, we conclude that both, topic information together with contexts of topic and argument, are important for the correct decision about a potential argument.
We observe similar effects in the three-classes scenario. Although in average different contexts for the recurrent model have a similar effect, we can clearly observe that taking topic information into account improves classification results by one score points. The combination of transfer learning for context and topic information again outperforms all other approaches by far. At the same time, the pre-trained model without topic information achieves a macro-f1 score of 0.61 which is 3 points lower than with topic information.




\subsection{Topic Dependent Cross-Topic Results}
As was shown in the previous subsection, argument classification produces satisfying results, especially if topic information and contexts are taken into account. 
In this set of experiments we evaluate the ability of different models to classify dependent on the topic. Therefore, a sentence may be considered to be an argument for one topic but be non argumentative for another. We argue that this is important, especially if text corpora are large, to filter out argumentative candidates which are arguments for different topics.
To evaluate the models ability to perform well in topic dependent classification we extend our dataset and change the experimental setting. For each topic we select a number of related terms. These are words which come from a similar context as a topic but it is very unlikely that the topic's argument are valid arguments for them. The list of related terms for each topic is provided in Table \ref{tab:topics}. For 50\% of argumentative sentences selected randomly from the test set, we replace the topic by one of the related terms of the topic and change the sentence label in the test set to non-argumentative. Therefore, to perform well on this task, a model should be able to recognize argumentative sentences in the context of the topic. To train for this task we correspondingly augment the training data. We keep the original training data and additionally select 50\% of argumentative sentences from the training set, select one of the related terms as topic, label them as non-argumentative and insert them into the training set. 
For this task we compare our model, which performed best on the original cross-topic task and compare it with the state-of-the-art approach \BiCLSTM. We also include the same models without topic information to see, whether topic information is still helpful or if the models get confused instead. 

The results for topic dependent classification are presented in Table \ref{tab:topic_dependent}. For the two-classes problem we observe a massive performance drop of ten points in macro-f1 score for the \BiCLSTM model. Nonetheless, the model still makes use of topic information and outperforms the standard \BiLSTM by two macro-f1 score points. Our approach \BertTI is more robust, the performance falls by moderate four score points and the gap to the counterpart model without topic information is incredible 17 score points large. We observe a similar behaviour in the three-classes scenario. Our \BertTI approach achieves the same average score as in the original cross topic task. In contrast the performance of the \BiCLSTM model drops by 11 score points and it even performs worse than the same model without topic information on this more complex task.
Thus we conclude that unlike previous models our approaches are indeed able to grasp the context of the argument and topic and are able to relate them with each other. 


\begin{table*}[]
\begin{center}
\begin{tabular}{|c|c|c|c|c|c|c|c|c|c|}
\hline
\multicolumn{1}{|c}{Topic} & \multicolumn{5}{|c|}{Related terms} \\
\hline
\bf abortion & euthanasia & teenage pregnancy & family & medical procedure & rape \\
\bf cloning & biology & species & religion & organ donation & modified food \\
\bf death penalty & politics& ethic& prison& homicide& sentence \\
\bf gun control & safety& school shooting& robbery& regulation& police state \\
\bf marijuana legalization & drugs & medicine & relaxation & freedom & liberty \\
\bf minimum wage & social justice& slavery& automation& economic crisis& stagnation \\
\bf nuclear energy & environment& employment& industry& pollution& climate change \\
\bf school uniforms & equality& social justice& individualism& clothing& mobbing \\
\hline
\end{tabular}
\end{center}
\caption{\label{tab:topics} Related terms for each topic}
\end{table*}

\begin{table*}[t]
\begin{center}
\begin{tabular}{|c|c|c|c|c|c|c|c|c|c|c|}
\cline{2-10}
\multicolumn{0}{c}{} & \multicolumn{1}{|c}{Method} & \multicolumn{8}{|c|}{Topics} \\
\cline{2-11} \multicolumn{1}{c}{} & \multicolumn{1}{|c|}{} & \multicolumn{1}{c|}{Abortion} & \multicolumn{1}{c|}{Cloning} &\multicolumn{1}{c|}{Death penalty} & \multicolumn{1}{c|}{Gun control }& \multicolumn{1}{c|}{Marij. legal.} & \multicolumn{1}{c|}{Min. wage} & \multicolumn{1}{c|}{Nucl. energy} &  \multicolumn{1}{c|}{School unif.} & \multicolumn{1}{c|}{$\diameter$}\\ \hline
 \parbox[t]{2mm}{\multirow{4}{*}{\rotatebox[origin=c]{90}{two-classes }}} 
& \BiLSTM & 0.57 &	0.59  & 0.53 &	0.59 &	0.62 &	0.62 &	0.59 &	0.57 &	0.58 \\
& \BiCLSTM & 0.62 &	0.72 & 0.46 & 0.46 &	0.76 &	0.60 &	0.69 &	0.45 &	0.60\\
& \Bert  & 0.56 &	0.63 &	0.60 &		0.62 &	0.61 &	0.55 &	0.60 &	0.53 &	0.59\\
& \BertTI & \bf0.68 & \bf0.77	&\bf 0.78 & \bf0.79 &\bf	0.82 &\bf	0.85 	 &\bf0.79 &	\bf0.58&	\bf 0.76\\
 \hline
 \parbox[t]{2mm}{\multirow{4}{*}{\rotatebox[origin=c]{90}{three-classes}}} 
& \BiLSTM & 0.39 &	0.39  & 0.37 &	0.36 &	0.39 &	0.42 &	0.40 &	0.39 &	0.39 \\
& \BiCLSTM & \bf 0.46 &	0.34 & 0.29 & 0.35 &	0.42 &	0.29 &	0.47 &	0.30 &	0.36\\
& \Bert  & 0.42 &	0.50 &	0.42 &	0.42 &	0.48 &	0.51 &	0.50 &	0.49 &	0.47\\
& \BertTI & 0.44 &	\bf 0.60 & \bf 0.52 & \bf 0.49 &	\bf 0.61 &	\bf 0.65 & \bf0.62&	 \bf0.55&	\bf0.56\\
\hline
\end{tabular}
\end{center}
\caption{\label{tab:topic_dependent} Topic dependent cross-topic  classification results}
\end{table*}

%% file: conclusion.tex
\section{Conclusion}
In this paper, we introduce a new approach for argument mining which takes a topic of the potential argument into account.
We hypothesize that considering information about the topic of a potential argument and their contexts should lead to better argument recognition.
We present multiple ways to include topic and contexts into the argument mining process. Precisely, we show how contexts from word embeddings, Knowledge Graph embeddings and models pre-trained on other tasks can be integrated into our approach. 
Our experimental results clearly show that considering topics in the decision process leads to better results in almost all considered cases. Especially our approach with topic information in connection with context from pre-trained models improves state-of-the-art approach by far in the real-world scenario. We also show that in contrast to current state-of-the-art methods, our approach is robust and able to perfectly grasp the  context of topic and potential argument. For future work we plan to focus more on Knowledge Graphs and other external context sources. In detail, we want to use information gathered from knowledge graphs not only for topics but also on the argument side. We also plan to investigate different Knowledge Graph embedding techniques and combine different Knowledge Graphs in the same model. For instance, a combination of fact based knowledge graphs like DBPedia \cite{dbpedia-swj} and Wikidata \cite{Vrandecic:2014:WFC:2661061.2629489} with knowledge graphs like WordNet \cite{Miller:1995:WLD:219717.219748} and FrameNet \cite{Baker:1998:BFP:980451.980860, Baker:1998:BFP:980845.980860} which focus on lexical similarities could further increase the representation quality of the context. Additional datasets with topic information about more topics will also deepen our understanding of the interplay between context and arguments and potentially further increase the performance of the argumentation models.
